 \documentclass[pmlr,twocolumn,10pt]{jmlr} % W&CP article

\usepackage{booktabs}
\usepackage{siunitx}

\newcommand{\comment}[1]{\ignorespaces}

\theorembodyfont{\upshape}
\theoremheaderfont{\scshape}
\theorempostheader{:}
\theoremsep{\newline}

\usepackage{booktabs, multirow}

\jmlrvolume{LEAVE UNSET}
\jmlryear{2023}
\jmlrsubmitted{LEAVE UNSET}
\jmlrpublished{LEAVE UNSET}
\jmlrworkshop{Machine Learning for Health (ML4H) 2023} % W&CP title

\title{Surpassing GPT-4 Medical Coding with a Two-Stage Approach}

\author{%
\Name{Zhichao Yang}
\Email{zhichaoyang@umass.edu}\\
\addr University of Massachusetts, Amherst, MA, USA
\AND
\Name{Sanjit Singh Batra} 
\Email{sanjit.batra@optum.com}\\
\Name{Joel Stremmel}
\Email{joel\_stremmel@optum.com}\\
\Name{Eran Halperin} 
\Email{eran.halperin@uhg.com}\\
\addr Optum AI, Minnetonka, MN, USA
}

\begin{document}

\maketitle

\begin{abstract}
Recent advances in large language models (LLMs) show potential for clinical applications, such as clinical decision support and trial recommendations. However, the GPT-4 LLM predicts an excessive number of ICD codes for medical coding tasks, leading to high recall but low precision. To tackle this challenge, we introduce LLM-codex, a two-stage approach to predict ICD codes that first generates evidence proposals using an LLM and then employs an LSTM-based verification stage. The LSTM learns from both the LLM's high recall and human expert's high precision, using a custom loss function. 
Our model is the only approach that simultaneously achieves state-of-the-art results in medical coding accuracy, accuracy on rare codes, and sentence-level evidence identification to support coding decisions without training on human-annotated evidence according to experiments on the MIMIC dataset.  

\end{abstract}
\begin{keywords}
Natural Language Processing, Large Language Models, Generative Models, Semi Supervised Learning, Explainability, Interpretability
\end{keywords}

\section{Introduction}
\label{sec:intro}

Clinical text encompasses a vast array of essential information that extends beyond the structured data fields obtained from electronic health records (EHRs) \citep{Zweigenbaum2007FrontiersOB,Uzuner2010ExtractingMI,Wang2018ClinicalIE,Yao2022AutomatedIO,pmlr-v193-li22a,Jiang2023HealthSL}. A critical task in EHR analysis is the assignment of International Classification of Diseases (ICD) codes \citep{Larkey1996Combining}, which entails attributing zero, one, or multiple ICD codes to a given note.
 
Computational methods have been employed to automate the task of ICD coding. Ideally, such computational methods should overcome the following challenges: (1) The first challenge is the scarcity of training data since labeling EHRs is an expensive process \citep{Wei2018ClinicalTA, Willemink2020PreparingMI}, often resulting in a scarcity of sufficient training data. (2) The second challenge is achieving high precision and recall for all ICD codes, including rare ones, as they may hold equal clinical importance for patients as common codes \citep{Atutxa2019InterpretableDL, Dong2021RareDI}. (3) The third challenge is explainability since it is crucial in the medical field to ensure trust in the classifier's decisions. Consequently, computational methods should be capable of providing sentence-level evidence to support their coding decisions.

Unfortunately, existing computational methods for medical coding fail to address all three critical issues concurrently. In particular, state-of-the-art medical coding models are unable to provide \textit{sentence-level} evidence for their coding decisions due to their black-box nature \citep{yuan-etal-2022-code, jain-wallace-2019-attention}. While some models do offer such sentence-level evidence, they necessitate training on annotated evidence, which requires substantial human annotation costs \citep{cheng-etal-2023-mdace}.

Recent studies have demonstrated that large language models (LLMs) can serve as effective few-shot learners when training examples are limited \citep{Zhao2021CalibrateBU, min-etal-2022-metaicl, chen-etal-2022-improving}. Furthermore, LLMs can be directly prompted for evidence to support their medical coding decisions, making them well-suited for this task \citep{agrawal-etal-2022-large}. However, we observe that state-of-the-art LLMs, such as GPT-4, exhibit low precision in medical coding tasks, as depicted in Figure~\ref{fig:initial_result}. As a result, there is currently no method that effectively addresses all three of these challenges simultaneously. 

\begin{figure}[htbp]
\floatconts
  {fig:initial_result}
  {\caption{An initial experiment on the accuracy of ICD coding compared between few-shot ICL GPT4 \citep{OpenAI2023GPT4TR} and fine-tuned MSMN}}
  {\includegraphics[width=0.99\linewidth]{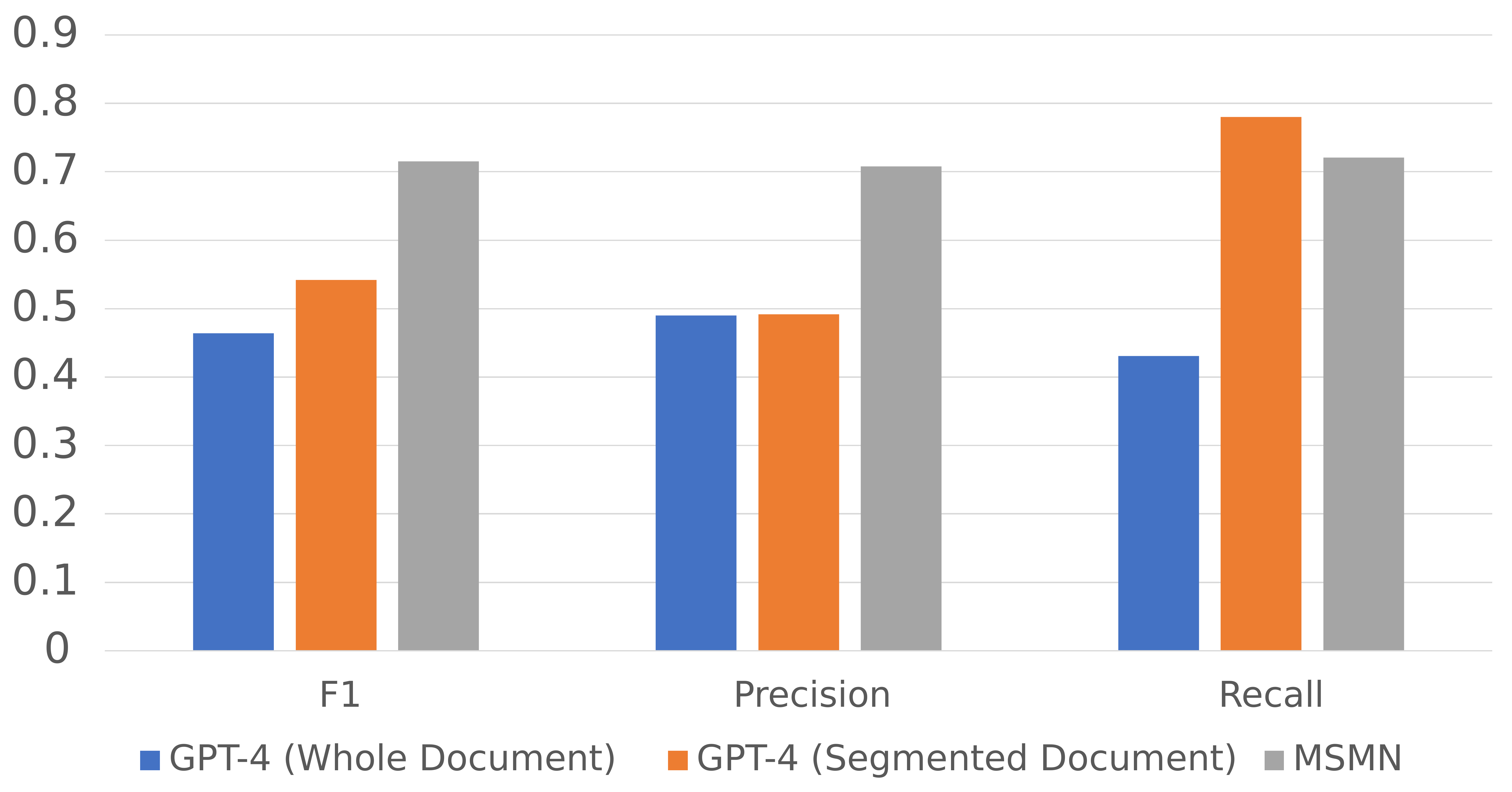}}
\end{figure}

In this paper, we propose a two-stage approach, \textbf{LLM-codex}, that addresses all three challenges simultaneously. This approach attains state-of-the-art medical coding accuracy even with limited training data and rare codes. Additionally, LLM-codex furnishes precise sentence-level evidence for coding decisions without necessitating training on annotated evidence. 

\begin{figure*}[htbp]
\floatconts
  {fig:pipeline}
  {\caption{An illustration of LLM-codex where we use an LLM to extract code-evidence pairs and then verify them with a \textit{Verifier} model. The examples are artificial and for demonstration only.}}
  {\includegraphics[width=0.99\linewidth]{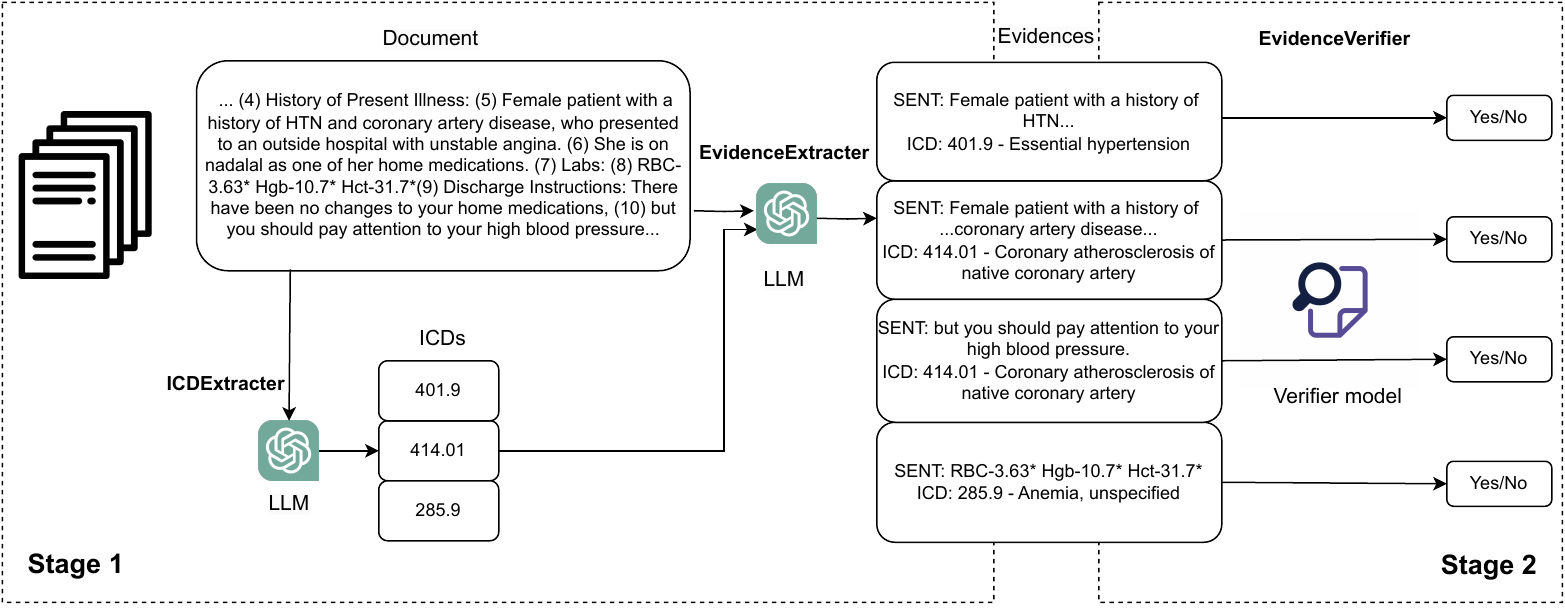}}
\end{figure*}

LLM-codex is a two-stage approach consisting of an LLM in the first stage and a \textit{Verifier} model in the second stage. 
In the first stage, we segment long EHRs into smaller segments and feed each segment into the LLM. While this strategy substantially improves recall, it leads to lower precision due to the over-prediction of ICD codes. Consequently, in the second stage, we introduce an additional filter—a \textit{Verifier} model—which verifies the predicted ICD codes \citep{zaidan-etal-2007-using}. 
Our Verifier model is an LSTM trained with a custom loss function leveraging dual labels: the LLM-assigned ICD code as a sentence-level, silver-label (high recall), and the expert-assigned ICD code as the document-level, gold-label (high precision). The Verifier is designed to assign scores to each sentence based on its ability to predict the corresponding ICD code.

Incorporating the LLM in the first stage and the Verifier model in the second stage, LLM-Codex attains a substantial improvement of over 10\% in F1 score for rare codes relative to state-of-the-art medical coding models. Furthermore, it exhibits about a 5\% increase in F1 score on limited training data. Additionally, without requiring training on annotated evidence, LLM-Codex boosts evidence accuracy by over 10\% when compared to the top-performing sentence-level evidence model for coding decisions.

As a result, LLM-Codex presents a comprehensive solution that tackles all three aforementioned issues concurrently, positioning itself as a promising framework for medical coding. We believe LLM codex can potentially be used on classification tasks beyond the medical domain that require providing supporting evidence for classification decision \citep{samek2017explainable}.

\section{Related work}
\label{sec:related}

Automated ICD coding employs natural language processing (NLP) models to predict expert-labeled ICD codes using EHRs as input. This problem has traditionally been formulated as a multi-label classification task. Early approaches, such as CAML \citep{mullenbach-etal-2018-explainable}, utilized a convolutional neural network to encode medical documents, followed by a label-wise attention mechanism to focus on the labeled ICD codes of the input notes during training. More recently, state-of-the-art methods have incorporated various techniques, such as incorporating synonyms of clinical concepts \citep{yuan-etal-2022-code, yang-etal-2022-knowledge-injected}, exploring the discourse structure within EHRs \citep{zhang-etal-2022-automatic}, and utilizing data augmentation \citep{falis-etal-2022-horses} to enhance performance. Additionally, advancements in the field have emerged from exploring alternative architectures, such as pretrained bidirectional language models \citep{huang-etal-2022-plm, michalopoulos-etal-2022-icdbigbird} and pretrained autoregressive language models combined with prompts \citep{Yang2022MultilabelFI}. In this paper, we propose a novel method, LLM-codex, to address the limitations of existing methods in automated ICD coding by leveraging a two-stage approach that significantly improves performance on rare coding labels.

The application of LLMs to unstructured clinical data has been a major focus of recent research \citep{jimenez-gutierrez-etal-2022-thinking,Zhou2022CancerBERTAC,McInerney2023CHiLLZC}. For instance, \citet{agrawal-etal-2022-large} demonstrated that LLMs can effectively extract information from clinical text, even without training on clinical data. Likewise, \citet{meoni-etal-2023-large} emphasized the potential of LLMs for information extraction tasks in the clinical domain, particularly when data is scarce due to confidentiality concerns arising from stringent privacy regulations that protect sensitive patient information. However, recent studies have found that LLMs struggle to extract information when tasks necessitate accessing relevant information within lengthy contexts \citep{Liu2023LostIT}. We address this challenge by segmenting long documents to enhance their flow and readability.

To elucidate the reasons for assigning an ICD code to a document, previous research has primarily relied on attribution maps, derived either from the salience of individual words or the attention weights of specific tokens \citep{mullenbach-etal-2018-explainable,pmlr-v126-lovelace20a,Dong2020ExplainableAC,liu-etal-2021-effective,kim-etal-2022-current,wang-etal-2022-novel,nguyen-etal-2023-two}. However, these attribution maps exhibit limited explanation accuracy \citep{Sinha2021PerturbingIF,Ivankay2022FoolingEI}. \citet{ivankay-etal-2023-dare} observed that when minor perturbations (modifications to a single task-irrelevant phrase or sentence) were introduced to a medical document, many words with initially positive attributions shifted to negative values, despite the code prediction remaining accurate. This issue resonates with the findings of \citet{jain-wallace-2019-attention}, who argued that attention-based explanations might not provide a complete understanding of model decisions. Their research demonstrated that attention weights can be easily manipulated without significantly affecting model predictions, that the same model with different attention weights could produce identical predictions, and that attention weights might remain unchanged even when input perturbations change the model's output. These findings, along with the limitations of attribution maps, emphasize the need for more reliable interpretability methods in ICD coding and other NLP tasks. 
In this paper, we address these challenges by proposing LLM-codex, which identifies the most relevant sentence from a long document when predicting an ICD code and subsequently verifies this sentence to produce a final prediction. This strategy is inspired by previous research demonstrating that the verification of LLMs can improve their output \citep{Weng2022LargeLM}.

\section{Datasets}
\label{sec:data}

We utilized several datasets to evaluate the model's coding performance and explainability.

\textbf{MIMIC-III \textit{common}}: MIMIC-III \citep{Johnson2016MIMICIIIAF} is a publicly accessible dataset containing discharge summary documents from an Intensive Care Unit (ICU), with each document associated with ICD codes labeled by medical coding experts. In line with prior work \citep{mullenbach-etal-2018-explainable}, we filtered the dataset to retain instances featuring at least one of the top 50 most frequent ICD codes. This results in 8,067 training instances and 1,729 test instances based on the canonical data splits from \citet{mullenbach-etal-2018-explainable}.

\textbf{MIMIC-III \textit{few-shot}}: To assess the model's performance under limited training data conditions, we randomly selected about one-eighth of the instances from the training data. This resulted in 1,000 training instances and 1,729 test instances. This subset comprises the top 50 most frequent ICD codes and $\sim14$ training instances per label (shot) on average, adhering to the few-shot criteria. 

\textbf{MIMIC-III \textit{rare}}: To assess the model's performance on predicting rare disease codes, which could be of equal importance as common disease codes, for a given patient, we built a rare code dataset using MIMIC-III. We collected rare diseases defined by medical experts \citep{Pavan2017ClinicalPG, NguengangWakap2019EstimatingCP}, and followed the pre-processing steps described in \citet{yang-etal-2022-knowledge-injected}. This resulted in $\sim5$ training instances per label (shot) on average. 

\textbf{MDACE Profee}: For evaluating the model's explainability, we used the code evidence dataset from \citet{cheng-etal-2023-mdace}. Expert annotators labeled a short text span for each ICD code assigned, indicating the rationale behind the assignment. The MIMIC-III dataset was annotated under professional fee billing guidelines, resulting in the \textit{MDACE Profee} datasets. We subsequently mapped each annotated text span to a sentence, serving as evidence for evaluation purposes. There are 172 sentence-ICD pairs in the evaluation dataset.

\section{Methods}
\label{sec:method}

\subsection{Task formulation}
ICD coding is typically formulated as a multi-label classification task, wherein the objective is to assign a binary label $y_{c,k} \in \{0, 1\}$ for each ICD code $c$ in the label space $Y$, given thousands of words from an input EHR document k. A label of 1 indicates that a medical document is positive for a specific ICD code. Candidate ICD codes can be described using a short code description phrase in free text, such as the description ``essential hypertension." corresponding to the ICD code 401.9. In addition to assigning the correct code, the goal is to also extract sentence-level evidence $m$ from the document for each $c$ to explain the model's decision.

To address these two tasks, we first employed an LLM to identify sentence-level evidence for all candidate ICD codes (\sectionref{sec:llmcode} and \sectionref{sec:llmevidence}). Subsequently, we used the ICD codes predicted by the LLM as \textit{silver} labels to train a \textit{Verifier} model that verifies whether the sentence-level evidence is accurate for the given ICD (\sectionref{sec:rm}).

\subsection{Stage 1a: Extracting document-level ICD codes using an LLM}
\label{sec:llmcode}

Utilizing an LLM such as GPT-4 \citep{OpenAI2023GPT4TR} with in-context learning (ICL) necessitates the specification of:

a) A template for providing input documents via the prompt;

b) An LLM to execute the prompt and generate output text;

c) A parser to convert the output text into the task-specific output space.

Thus, we first used the LLM to extract ICD codes using ICL. To achieve this, we carefully designed our prompt templates using a single ICL example. As depicted in Example \ref{example:icdcode_promptoriginal}, the template instructed the LLM to emulate a proficient clinical coding expert and assign a list of ICD codes to the given document. In order to effectively manage long documents, we first split it into multiple segments containing an equal number of sentences and passed each segment individually to the LLM. The LLM then predicted which of the candidate ICD codes are present in each segment, in the form of free text which was then parsed into a Python list of predicted ICD codes. Finally, we aggregated the ICD code predictions obtained from each EHR segment to generate the LLM's document-level ICD code predictions.

\subsection{Stage 1b: Extracting sentence-level ICD code evidence using an LLM}
\label{sec:llmevidence}
Given the document-level ICD code predictions, we used the LLM to identify sentence-level evidences for each predicted ICD code. Similar to the document-level ICD code prediction, we split the EHR into multiple segments containing an equal number of sentences. As demonstrated in Example \ref{example:evidence}, the template guided the LLM to emulate an evidence extraction expert by scanning each sentence in the segment and assigning one or more of the predicted document-level ICD codes to each sentence.

The LLM's output was subsequently parsed and aggregated across the segments within an EHR\footnote{\url{https://api.python.langchain.com/en/latest/chains/langchain.chains.llm.LLMChain.html}} to generate a Python list of tuples, wherein each tuple comprised an ICD code and its corresponding evidence sentence index. 

\subsection{Stage 2: Verifying sentence-level evidence using a \textit{Verifier} model}
\label{sec:rm}
Upon extracting pairs of predicted ICD codes $c$ and evidence sentences $m$ using the LLM, we verified the relationship between the pairs with the help of a \textit{Verifier} model \citep{zaidan-etal-2007-using}. The \textit{Verifier} model assessed the accuracy of a silver label (which consists of an ICD code and its corresponding evidence sentence index) predicted by the LLM. To accomplish this, LLM-codex first split the document into sentences and subsequently ranked these sentences based on their relevance to predicting the document-level \textit{gold} labels $y_c$ on each ICD code $c$. Additionally, it incorporated supervision from LLM-assigned \textit{silver} labels $y'_c$ for each sentence, during the ranking process.

We denote the set of sentence-level evidences corresponding to the \textit{silver} labels obtained by the LLM for the $k$-th document as:
\begin{equation}\label{eq:eq1m}
m_{k} = [m_{k,1}, ..., m_{k,j}, ..., m_{k,S_k}]
\end{equation}

where $S_k$ is the total number of sentence-level evidences identified by the LLM in document $k$.

We then used the \textit{Verifier} model iteratively across each predicted document-level ICD code $c$, to verify which of the predicted sentence-level evidences truly correspond to $c$. 
We therefore represented the predicted \textit{silver} labels for the $k$-th document as $x_{c, k}$ where:
\begin{equation}\label{eq:eq1x}
x_{c, k} = [(m_{k,1}, y'_{c,k,1}), ..., (m_{k,{S_k}}, y'_{c,k,S_k})]
\end{equation}

where $y'_{c,k} \in \mathbb{R}^{S_k}$ and $y'_{c,k,j}$ was 1 if and only if, $m_{k,j}$ was predicted to have evidence for $c$ in the \textit{silver} labels. 

The \textit{Verifier} model consists of a text encoder $TE$ which transforms a sentence-level evidence $m_{k, j}$ into its latent representation, $h^m_j$, using the following:
\begin{equation}\label{eq:eq2-1}
h^m_j = TE(m_{k, j})
\end{equation}

We followed MSMN \citep{yuan-etal-2022-code} and used an LSTM \citep{Hochreiter1997LongSM} as text encoder $TE$. It also transforms the short ICD code description, $c_{description}$, of code $c$, into its latent representation, $h^c$, using the following:
\begin{equation}\label{eq:eq2-2}
h^c = TE(c_{description})
\end{equation}

The per-label-attention $AT$ then combines the latent representations computed above to obtain label-specific logits $z_{k,j}$ \citep{mullenbach-etal-2018-explainable, liu-etal-2021-effective, yuan-etal-2022-code}:

\begin{equation}\label{eq:eq2-4}
z_{k,j} = AT(h^m_j, h^c)
\end{equation}

where $z_{k,j} \in \mathbb{R}^2$ because each label takes on one of two binary values in the ICD coding task.

The loss function corresponding to the \textit{Verifier} model was designed to consist of two terms, $l_{gold}$ and $l_{silver}$. Inspired by \citet{clark-gardner-2018-simple} and \citet{min-etal-2019-discrete}, the first term, $l_{gold}$, can be written as the weighted sum of losses corresponding to each sentence-level evidence, as follows: 

\begin{equation}\label{eq:eq5}
l_{gold} = \sum_{j=1}^{S_k} w_{k,j} l_{k,j}
\end{equation}

where, $l_{k,j}$ is the cross-entropy loss computed using $z_{k,j}$ and the document-level \textit{gold} label $y_{c,k}$ corresponds to the ICD code $c$ on document $k$. 

In order to compute the weight $w_{k,j}$, we first performed a maximum operation over the two dimensions of $z_{k,j}$ and then normalized them across $j$ using a softmax function. Therefore,
\begin{equation}\label{eq:eq4}
w_{k,j} = softmax(max(z_{k,j}))
\end{equation}

The second term in the loss function, $l_{silver}$ uses the \textit{silver} labels, $y'_k$, and can be written as: 

\begin{equation}\label{eq:eq6}
l_{silver} = \sum_{j=1}^{S_k} l'_{k,j}
\end{equation}

where $l'_{k,j}$ computes the cross-entropy loss between $y'_{c,k,j}$ and confidence score logits $z'_{k,j}$. To obtain $z'_{k,j}$ we first computed a maximum over the two dimensions of $z_{k,j}$: 

\begin{equation}\label{eq:eq7}
z'_{k,j} = max(z_{k,j})
\end{equation}

Finally, we trained the \textit{Verifier} model with the total loss for the $k$-th document, corresponding to ICD code $c$ as follows, 

\begin{equation}\label{eq:eq8}
L_{k,c} = l_{gold} + l_{silver}
\end{equation}

To make predictions for code $c$ in document $k$, we first select the sentence index $j$ with the highest weight $w_{k,j}$ among all candidate sentences $m_k$. If the $argmax$ over the two-dimensional $z_{k,j}$ corresponds to the positive label, we then output its corresponding value as the prediction score for the code $c$.

\subsection{Baselines for benchmarking}

\begin{enumerate}
    \item CAML \citep{mullenbach-etal-2018-explainable} uses a convolutional layer to extract features from an EHR and an attention mechanism to select the most relevant part of the EHR for predicting each ICD code.
    \item MSMN \citep{yuan-etal-2022-code} uses code description synonyms with multi-head attention and achieves state-of-the-art performance on the MIMIC-III \textit{common} task.
    \item EffectiveCAN with supervised attention \citep{cheng-etal-2023-mdace} employs a convolutional attention network to train on both document-level labels and evidence annotations using supervised attention. Their evidence annotations are generated by clinical coding experts, in contrast to our evidence (\textit{silver}) labels which are obtained from an LLM.
    \item Medalpaca \citep{han2023medalpaca} is a 13 billion parameter LLM trained to answer 1.5 million medical question. We replaced GPT-4 with this model to see how different LLMs performs.
\end{enumerate}

\section{Results}

\subsection{Predicting document-level common ICD codes with limited training data}

\begin{table}[htbp]
\floatconts
  {tab:fewshot}
  {\caption{Coding performance on MIMIC-III \textit{few-shot}. Mean and standard deviation over 20 experiments are shown.}}
  {
\footnotesize
  \begin{tabular}{lrrrr}\toprule
\multirow{2}{*}{Model} 
&\multicolumn{2}{c}{ROCAUC} &\multicolumn{2}{c}{F1} \\\cmidrule{2-5}
&MACRO &MICRO &MACRO &MICRO  \\\midrule
\multirow{2}{*}{CAML} &0.665 &0.729 &0.258 &0.364   \\
&$\pm0.003$ &$\pm0.004$ &$\pm0.007$ &$\pm0.014$ \\
\multirow{2}{*}{MSMN} &0.833 &0.874 &\textbf{0.489} &0.561  \\
&$\pm0.012$ &$\pm0.007$ &$\pm0.010$ &$\pm0.006$ \\
\multirow{2}{*}{\scriptsize{EffectiveCAN}} &0.802 &0.871 &0.434 &0.556 \\
&- &- &- &- \\
\multirow{2}{*}{Medalpaca} &0.435 &0.636 &0.189 &0.224 \\
&$\pm0.011$ &$\pm0.009$ &$\pm0.023$ &$\pm0.015$ \\
\multirow{2}{*}{LLM-codex} &\textbf{0.834} &\textbf{0.911} &0.468 &\textbf{0.611} \\
&$\pm0.006$ &$\pm0.005$ &$\pm0.017$ &$\pm0.015$ \\
LLM-codex  &0.511 &0.737 &0.169 &0.382  \\
/wo silver &$\pm0.011$ &$\pm0.004$ &$\pm0.003$ &$\pm0.011$ \\
\bottomrule
  \end{tabular}
  }
\end{table}

\label{sec:result}
First, we benchmarked LLM-codex on the medical coding task using the MIMIC-III few-shot dataset described in Section \ref{sec:data}. LLM-codex achieved a micro F1 of 0.611, which represents $\sim5$\% (absolute) improvement over existing approaches (\tableref{tab:fewshot}), and a micro ROCAUC of 0.911, $\sim3$\% (absolute) improvement over all existing methods (\tableref{tab:fewshot}). Similar performance improvements were observed for ICD-10 prediction with limited training data (\tableref{tab:fewshoticd10}).

We also found that removing the silver labels obtained using the LLM from LLM-codex's training process led to a significant decline in ICD coding prediction metrics, highlighting their crucial role in its performance.

We also found that different LLMs performed very differently, GPT-4 performed the best among all baselines while Medalpaca performed the worst in ROCAUC and F1 score. 

To investigate the impact of training data quantity on LLM-codex's performance, we benchmarked it on the MIMIC-III \textit{common} dataset with 3 different size of training data. When trained with all 8066 instances, LLM-codex performed on par with existing methods in terms of coding accuracy (\tableref{tab:vary}). When trained as few as 1000 and 500 instances, LLM-codex outperformed existing methods. This robustness highlights LLM-codex's potential with constrained data resources.

\subsection{Predicting document-level rare ICD codes}

\begin{table}[htbp]
\floatconts
  {tab:rare}
  {\caption{Coding performance on the MIMIC-III \textit{rare} }}
  {
\footnotesize
  \begin{tabular}{lrrrrr}\toprule
\multirow{2}{*}{Model} &\multicolumn{2}{c}{ROCAUC} &\multicolumn{2}{c}{F1} \\\cmidrule{2-5}
&MACRO &MICRO &MACRO &MICRO \\\midrule
\multirow{2}{*}{CAML } &0.574 &0.602 &0.072 &0.083 \\
&$\pm0.004$ &$\pm0.003$ &$\pm0.006$ &$\pm0.004$ \\
\multirow{2}{*}{MSMN } &0.755 &0.761 &0.169 &0.173 \\
&$\pm0.002$ &$\pm0.002$ &$\pm0.002$ &$\pm0.003$ \\
\multirow{2}{*}{LLM-codex} &\textbf{0.825} &\textbf{0.832} &\textbf{0.279} &\textbf{0.302} \\
&$\pm0.003$ &$\pm0.002$ &$\pm0.004$ &$\pm0.005$ \\
\bottomrule
  \end{tabular}
  }
\end{table}
To evaluate LLM-codex's performance on rare ICD codes, we assessed it using the MIMIC-III \textit{rare} dataset described in Section \ref{sec:data}.

We found that LLM-codex achieved an absolute improvement of $\sim12$\% in micro F1 and $\sim5$\% in micro ROCAUC compared to existing approaches (\tableref{tab:rare}).

These results further support the notion that LLMs are effective few-shot learners, capable of outperforming existing classification models fine-tuned for rare ICD code prediction \citep{lewis-etal-2020-pretrained, Taylor2022ClinicalPL, Shyr2023IdentifyingAE}.

\subsection{Ablation study of LLM-codex}

\begin{table}[htbp]
\floatconts
  {tab:ablation}
  {\caption{Ablation study of LLM-codex on MIMIC-III \textit{few-shot}. Micro scores are reported.}}
  {
  \small
\begin{tabular}{lrrrr}\toprule
Model &F1 &Precision &Recall \\\midrule
\multirow{2}{*}{Blackbox CAML } &0.365 &0.349 &0.383 \\
&$\pm0.014$ &$\pm0.005$ &$\pm0.009$ \\
\multirow{2}{*}{Blackbox MSMN } &0.561 &0.545 &0.581 \\
&$\pm0.006$ &$\pm0.010$ &$\pm0.019$ \\\midrule
\multirow{2}{*}{LLM-codex} &\textbf{0.611} &0.587 &0.638 \\
&$\pm0.015$ &$\pm0.015$ &$\pm0.015$ \\
LLM-codex (stage 1) &0.339 &\textbf{0.648} &0.230 \\
+ random forest &$\pm0.016$ &$\pm0.024$ &$\pm0.013$ \\
\multirow{2}{*}{LLM-codex (stage 1)} &0.493 &0.388 &0.674 \\
&$\pm0.010$ &$\pm0.011$ &$\pm0.011$ \\
\multirow{2}{*}{LLM-codex (stage 1a)} &0.360 &0.233 &\textbf{0.792} \\
&$\pm0.009$ &$\pm0.007$ &$\pm0.009$ \\
\bottomrule
\end{tabular}
  }
\end{table}

To better understand the factors contributing to LLM-codex's predictive performance, we compared it to three variants: one using the LLM only for ICD code extraction (Stage 1a), another without the \textit{Verifier} model (Stage 1), and a third where the \textit{Verifier} model was replaced by a random forest classifier. In the last variant, we counted the occurrence of evidence sentences per ICD code for the ICD code and evidence sentence index pairs extracted by the LLM and then used the occurrence matrix as features to train the random forest for ICD code verification.

We present the results on the MIMIC-III few-shot dataset in \tableref{tab:ablation} and make the following observations:

\begin{enumerate}
\item Implementing only Stage 1a of LLM-codex resulted in a significant decline in F1 score for ICD code prediction.
\item Including Stage 1b, which extracts sentence-level evidence for predicted ICD codes, improved the F1 score by $\sim13$\% (absolute) compared to LLM-codex with just Stage 1a.
\item Substituting the Verifier model with a random forest model led to a reduction in the F1 score by $\sim27$\% (absolute) compared to LLM-codex.
\end{enumerate}

In summary, both stages of LLM-codex significantly contributed to its ICD coding predictive performance.

\subsection{Ablation study of EHR segmentation on GPT-4}
We examined two distinct methodologies for prompting GPT-4 to perform ICD coding prediction: one involved presenting the entire document, while the other presented 10 equal-sized sentence segments of the document and aggregated the results across these segments. The latter approach (GPT4-seg) significantly increased recall (while maintaining comparable precision) compared to using the whole document as input (GPT4-doc) (\tableref{tab:gpt4}). This finding aligns with literature reports that LLMs face challenges in extracting information from the middle of long contexts \citep{Liu2023LostIT}. Despite this increase in recall, LLM-codex outperformed both methods in terms of F1 score on ICD code prediction.

\begin{table}[htbp]
\floatconts
  {tab:gpt4}
  {\caption{Ablation of EHR segmentation on MIMIC-III \textit{few-shot} ICD code prediction with GPT-4}}%
  {
\begin{tabular}{lrrrr}\toprule
Model &F1 &Precision &Recall \\\midrule
\multirow{2}{*}{GPT4-seg} &0.582 &0.482 &\textbf{0.730} \\
&$\pm0.010$ &$\pm0.011$ &$\pm0.011$ \\
\multirow{2}{*}{GPT4-doc} &0.484 &0.500 &0.471 \\
&$\pm0.015$ &$\pm0.009$ &$\pm0.010$ \\
\multirow{2}{*}{LLM-codex} &\textbf{0.611} &\textbf{0.587} &0.638 \\
&$\pm0.015$ &$\pm0.015$ &$\pm0.015$ \\
\bottomrule
\end{tabular}
  }
\end{table}

\subsection{Predicting sentence-level evidence for common ICD codes}

To assess LLM-codex's explainability capabilities, we utilized the MDACE Profee dataset \citet{cheng-etal-2023-mdace} outlined in Section \ref{sec:data}, which comprises ICD code evidence annotations created by professional medical coders. For each ICD code, LLM-codex provides a single sentence-level evidence if the predicted score exceeds a threshold optimized for that ICD code based on the F1 score of a validation dataset. LLM-codex selects the sentence-level evidence with the highest confidence score generated by the \textit{Verifier} model. We considered the evidence for each ICD code in an EHR as a true positive if a method captured at least one of its expert-annotated sentence-level evidences from the MDACE study \citep{glockner-etal-2020-think}.

Our findings indicated that LLM-codex yielded sentence-level evidences with the highest precision compared to existing methods like EffectiveCAN, which were trained on evidence annotations (\tableref{tab:evidence}). This result is consistent with prior literature that highlights LLMs as proficient medical evidence extractors \citep{McInerney2023CHiLLZC, Gero2023SelfVerificationIF}. While GPT4-seg exhibited the highest recall, a detailed analysis of its outputs uncovered an over-prediction of sentences as evidence for predicted ICD codes, leading to reduced precision (\tableref{tab:evidence}). As a result, LLM-codex surpasses existing methods in F1 score, striking a better balance between the precision and recall of its provided sentence-level evidences.

\begin{table}[htbp]
\floatconts
  {tab:evidence}
  {\caption{Benchmarking sentence-level evidence extraction in the MDACE Profee evaluation dataset}}
  {
\begin{tabular}{lrrrr}\toprule
Model &F1 &Precision &Recall \\\midrule
EffectiveCAN &0.542 &0.408 &0.806 \\
GPT4-seg &0.123 &0.066 &\textbf{0.944} \\
GPT4-doc &0.675 &0.596 &0.778 \\
LLM-codex &\textbf{0.713} &\textbf{0.608} &0.861 \\
\bottomrule
\end{tabular}
  }
\end{table}

\subsection{Error case analysis}
LLM-codex tends to overlook some ICD codes when the length of the sentence is long, as shown in row 2 and 3 in the \tableref{tab:casestudy}. Lengthier sentences typically have more ICD codes to assign, which can reduce GPT-4’s accuracy. Additionally, LLM-codex tends to assign ICD V-codes excessively. V-codes are used to indicate non-diagnostic information, such as preventive services, routine check-ups, and administrative encounters. Since fee-for-service payment systems do not incentivize coding V-codes, they are rarely utilized \citep{Torres2017ICDSC,Guo2020InternationalCO}. Hence, the ground truth may be under-labeled. The overprediction of ICD V-codes using GPT-4 may further support this line of research in healthcare.

\section{Discussion}

In this paper, we present LLM-codex, a two-stage model that leverages LLMs for predicting document-level ICD codes and their corresponding sentence-level evidence. Our results show that LLM-codex significantly outperforms prior state-of-the-art models in predicting common document-level ICD codes, particularly when faced with limited training data. Additionally, LLM-codex demonstrates superior performance in predicting document-level rare ICD codes. When a single sentence-level evidence suffices to justify predicted ICD codes, LLM-codex notably achieves higher precision compared to existing approaches.

Our work has several limitations.
First,
We found that when comprehensively extracting all available sentence-level evidence for a predicted ICD code is essential, GPT-4 with segmentation outperforms LLM-codex (\tableref{tab:evi_evalall}). This is due to LLM-codex's current constraint of generating only one sentence-level evidence per predicted ICD code. To boost LLM-codex's recall, one could increase the number of evidence sentences returned by the \textit{Verifier} model for each predicted ICD code. Although the impact on its precision remains unclear, exploring this modification could be part of future work. Moreover, incorporating explainability methods like the Masked Sampling Procedure (MSP) \citep{pmlr-v193-stremmel22a} into the \textit{Verifier} model could further enhance LLM-codex's explainability by more comprehensively identifying sentence-level evidence for each predicted ICD code.
Second, case studies show limited accuracy in long sentences, as GPT-4 can be misled by many medical keywords in a sentence. Finally, LLM-codex requires GPT-4 during inference. We estimate it costs \$0.50 per discharge summary to run LLM-Codex on the MIMIC-III dataset with a latency of about 10 seconds per document. To reduce cost and latency, future work could distill GPT-4 ICD coding performance into other large language models such as Llama2.

LLM-codex constitutes a substantial advancement in ICD code prediction and explainability by accurately predicting ICD codes, even with limited training data and for rare codes, while providing sentence-level explanations for coding decisions—capabilities not concurrently demonstrated by existing approaches. We believe this versatile method has the potential to extend to various classification tasks with limited annotations which require explanations for model decisions, such as medication abuse detection \citep{Fleming2008ReportedLA,Kwon2023ODDAB} and social determinants of health identification \citep{davidsonetalScreening, Mitra2022AssociationsBN}, thus paving the way for promising future research.

% TODO: uncomment for final submission.
%\acks{We would like to thank Ardavan Saeedi, Hamid Hassanzadeh, Brian Hill, and Jagadish Venkataraman for many helpful discussions. We thank the anonymous reviewers and our colleagues at OptumLabs for their insightful feedback that helped improve the paper.}

\bibliography{anthology, jmlr-sample}

\newpage
\section{Appendix}\label{appendix}
\renewcommand\thefigure{A.\arabic{figure}}  
\renewcommand\thetable{A.\arabic{table}}
\setcounter{figure}{0}
\setcounter{table}{0}

\subsection{Implementation Details}
For the experiments in this study, the LLM we employ is GPT4-8k version 0314 \citep{OpenAI2023GPT4TR}.
It is accessed securely through the Azure OpenAI API under the responsible use requirement\footnote{https://physionet.org/news/post/415}. We set the sampling temperature to 0.1 and truncate the EHRs to satisfy the 8k token constraint. Additionally, we define and evaluate the number of candidate codes $N_c$ as 50; in theory, $N_c$ could vary depending on the specific application. The LSTM architecture of our Verifier is the same as MSMN. Detailed hyperparameters are reported in \tableref{tab:hyperparameters}.

\subsection{Empirical results on the number of segments to split}
In \tableref{tab:gpt4}, we showed that breaking down the input patient record into multiple equal-size sentence segments and then aggregating results across these segments increased the F1 of GPT-4. To find the best number of segments $segn$, we tuned the $segn$ as hyperparameter and observed the best F1 when $segn=10$ in \tableref{tab:segn}.

\subsection{Benchmarking comprehensive evidence extraction}

In \tableref{tab:evidence}, an ICD code in an EHR is considered positively predicted by a method if it predicts at least one expert-annotated sentence-level evidence from the MDACE Profee dataset corresponding to that ICD code. However, for some applications, it may be beneficial to assess the differences between methods that capture more than one sentence-level evidence for a given ICD code in an EHR. Consequently, we introduce an evaluation of comprehensive sentence-level evidence extraction, where each expert-annotated sentence-level evidence for an ICD code in an EHR is treated as an individual data point, allowing a method to predict multiple positives for that ICD code in the EHR.

We observe that while LLM-codex maintains superior precision compared to existing methods, it exhibits the lowest recall (\tableref{tab:evi_evalall}). This is due to the fact that the average number of gold evidence labels per ICD code in the MDACE Profee dataset is 3, while LLM-codex outputs at most one sentence-level evidence for each ICD code (providing exactly one sentence-level evidence for all ICD codes whose predictions exceed a threshold optimized for the ICD code based on the F1 score of a validation dataset). Notably, we find that GPT4-seg achieves the highest recall, consistent with \tableref{tab:evidence}, but has low precision. Thus, the method demonstrating the optimal balance between precision and recall, and achieving the highest F1 score, is GPT4-doc, which outperforms EffectiveCAN in terms of F1, precision, and recall.

\subsection{ICD-10 accuracy evaluation}
We also tested coding accuracy on ICD-10 codes. We followed \citet{Nguyen2023MimicIVICDAN} and filtered the dataset to include instances with at least one of the top 50 most frequent ICD-10 codes. We limited the number of training instances to 1000 and named this dataset MIMIC-IV \textit{few-shot}. The result is shown in \tableref{tab:fewshoticd10}.

\subsection{Disease-specific ablation study of LLM-codex}
We investigate whether LLM-codex's performance varies across individual ICD codes and which of its components are critical for its performance. In order to do so, we first locate mentions of ICD-9 codes with an NER tool MedCat \citep{Kavuluru2015AnEE}. We then evaluate ICD coding accuracy on codes that were not explicitly mentioned in the documents.

We observe that on anemia prediction, LLM-codex with stage 1 and 2 achieves an F1 score of 0.567, outperforming MSMN which only scores 0.252 F1 (Figure \ref{fig:casestudy}). Among the documents that had the hypertension code assigned, $\sim8$\% of the EHRs were missing mentions of hypertension. In comparison, $\sim55$\% of EHRs were missing mentions of anemia, thereby making the task of predicting anemia harder as it would require inference without it being explicitly mentioned. LLM-codex performs on par with MSMN in predicting hypertension. Furthermore, on the task of predicting anemia, LLM-codex achieves an AUPRC of, significantly outperforming MSMN's AUPRC of 0.208.

\begin{table*}[htbp]
\floatconts
  {tab:fewshoticd10}
  {\caption{Results on the MIMIC-IV \textit{few-shot} benchmark on ICD-10 codes}}%
  {
\begin{tabular}{lrrrrr}\toprule
\multirow{2}{*}{Model} 
&\multicolumn{2}{c}{ROCAUC} &\multicolumn{2}{c}{F1} \\\cmidrule{2-5}
&MACRO &MICRO &MACRO &MICRO \\\midrule
\multirow{2}{*}{MSMN } &\textbf{0.840} &0.883 &\textbf{0.508} &0.577 \\
&$\pm0.015$ &$\pm0.010$ &$\pm0.017$ &$\pm0.015$ \\
\multirow{2}{*}{LLM-codex } &0.837 &\textbf{0.906} &0.497 &\textbf{0.604} \\
&$\pm0.005$ &$\pm0.007$ &$\pm0.009$ &$\pm0.074$ \\
\bottomrule
  \end{tabular}
  }
\end{table*}

\begin{table*}[htbp]
\floatconts
  {tab:common}
  {\caption{Results on the MIMIC-III \textit{common} benchmark}}%
  {
  \begin{tabular}{lrrrrr}\toprule
\multirow{2}{*}{Model} 
&\multicolumn{2}{c}{ROCAUC} &\multicolumn{2}{c}{F1}   \\\cmidrule{2-5}
&MACRO &MICRO &MACRO &MICRO &  \\\midrule
\multirow{2}{*}{Blackbox CAML } &0.884 &0.916 &0.576 &0.633  \\
&- &- &- &- \\
\multirow{2}{*}{Blackbox MSMN } &0.928 &\textbf{0.947} &\textbf{0.683} &\textbf{0.725} \\
&$\pm0.004$ &$\pm0.003$ &$\pm0.007$ &$\pm0.008$ \\
\multirow{2}{*}{EffectiveCAN} &0.920 &0.945 &0.668 &0.717 \\
&- &- &- &- \\
\multirow{2}{*}{LLM-codex} &\textbf{0.929} &0.948 &0.674 &0.715  \\
&$\pm0.003$ &$\pm0.002$ &$\pm0.006$ &$\pm0.005$\\
\multirow{2}{*}{LLM-codex /wo silver label} &0.555 &0.762 &0.218 &0.435  \\
&$\pm0.004$ &$\pm0.003$ &$\pm0.007$ &$\pm0.004$ \\
\bottomrule
  \end{tabular}
  }
\end{table*}

\begin{table*}[htbp]
\floatconts
  {tab:vary}
  {\caption{Results on the MIMIC-III benchmark with varying number of training instances.}}
  {
\begin{tabular}{ll|rrr}\toprule
Train size (shot) &Model &F1 &Precision &Recall \\\midrule
8066 (all) &LLM-codex &0.715 &0.704 &0.726 \\
&MSMN &0.725 &0.713 &0.738 \\
1000 (14) &LLM-codex &0.611 &0.587 &0.638 \\
&MSMN &0.561 &0.545 &0.581 \\
500 (10) &LLM-codex &0.498 &0.475 &0.523 \\
&MSMN &0.413 &0.394 &0.434 \\
\bottomrule
\end{tabular}
  }
\end{table*}

\begin{table*}[htbp]
\floatconts
  {tab:casestudy}
  {\caption{Examples of qualitative evaluations on LLM-Codex}}
  {
\begin{tabular}{m{25em}|m{8em}|r}\toprule
Sentence &ICD code &Status \\\midrule
CT abdomen showed colitis. &556.8: Other ulcerative colitis &correct \\
\bottomrule
The patient is a 46 year old female with a history of hypertension, OSA, and depression who was transferred from [**Hospital1 **] after presenting to the ED there with 4 days of nausea, vomiting, diarrhea, and worsening jaundice. &401.9: Essential Hypertension &correct \\
\bottomrule
The patient is a 46 year old female with a history of hypertension, OSA, and depression who was transferred from [**Hospital1 **] after presenting to the ED there with 4 days of nausea, vomiting, diarrhea, and worsening jaundice. &782.4: Jaundice, unspecified, not of newborn &missed \\
\bottomrule
Patient is a 56 y/o M s/p Whipple resection + SMV reconstruction [**5-14**] for pancreatic ca with SMV thrombosis ([**Doctor Last Name **] and [**Doctor Last Name **]), with post-operative course marked by delayed gastric emptying, requiring NGT reinsertion on POD 7 until POD 11. &536.3: Gastroparesis &missed \\
\bottomrule
3 packs a day &V15.82: Personal history of tobacco use &overpredicted \\
\bottomrule
Post-operative course was characterized by fever &V45.81: Postsurgical aortocoronary bypass status &overpredicted \\
\bottomrule
\end{tabular}
  }
\end{table*}

\begin{table*}[htbp]
\floatconts
  {tab:hyperparameters}
  {\caption{Hyperparameters used to train \textit{Verifier} model}}
  {
\begin{tabular}{lrr}\toprule
Parameters &Value \\\midrule
Emb. dim. &100 \\
Emb. dropout &0.2 \\
LSTM Layer &1 \\
LSTM hidden dim. &512 \\
LSTM output dim. &512 \\
Rep. dropout &0.2 \\
learning rate &5e-4 \\
Batch size &1 doc \\
Weight decay &0.02 \\
\bottomrule
\end{tabular}
  }
\end{table*}

\begin{table*}[htbp]
\floatconts
  {tab:segn}
  {\caption{Best $segn$ for GPT-4 model on MIMIC-III \textit{few-shot} ICD code prediction task}}
  {
\begin{tabular}{lrrrr}\toprule
Model &F1 &Precision &Recall \\\midrule
\multirow{2}{*}{GPT4 (segn=50)} &0.346 &0.225 &\textbf{0.749} \\
&±0.009 &±0.008 &±0.011 \\
\multirow{2}{*}{GPT4 (segn=25)} &0.511 &0.389 &0.744 \\
&±0.010 &±0.011 &±0.011 \\
\multirow{2}{*}{GPT4 (segn=10)} &\textbf{0.582} &0.482 &0.73 \\
&±0.010 &±0.011 &±0.011 \\
\multirow{2}{*}{GPT4 (segn=1)} &0.484 &\textbf{0.5} &0.471 \\
&±0.015 &±0.009 &±0.010 \\
\bottomrule
\end{tabular}
  }
\end{table*}

\begin{table*}[htbp]
\floatconts
  {tab:evi_evalall}
  {\caption{Benchmarking sentence-level \textit{comprehensive} evidence extraction in the MDACE Profee evaluation dataset}}%
  {
\begin{tabular}{lrrrr}\toprule
Model &F1 &Precision &Recall \\\midrule
EffectiveCAN &0.480 &0.393 &0.616 \\
GPT4-seg &0.103 &0.055 &\textbf{0.860} \\
GPT4-doc &\textbf{0.550} &0.455 &0.698 \\
LLM-codex &0.453 &\textbf{0.608} &0.360 \\
\bottomrule
\end{tabular}
  }
\end{table*}

\begin{figure*}[htbp]
\floatconts
  {fig:subfigex}
  {\caption{Bin plot on the location of evidences, where x-axis is the sentence position from the start of document and y-axis is the occurrence density of each bin.}}
  {%
    \subfigure[Human]{\label{fig:abc_gold}%
      \includegraphics[width=0.24\linewidth]{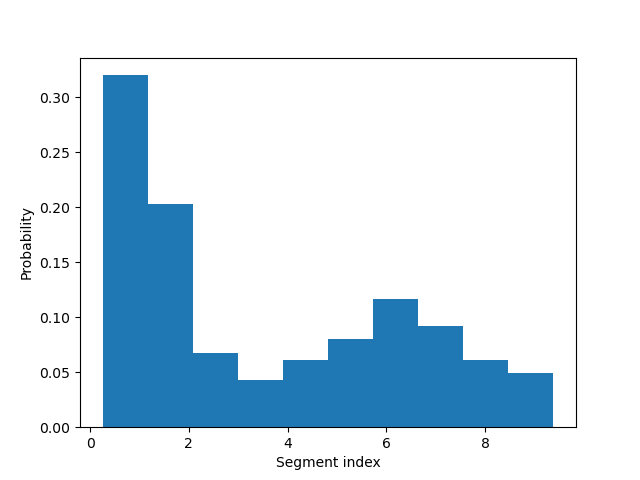}}%
    \subfigure[LLM-codex]{\label{fig:abc_modelc}%
      \includegraphics[width=0.24\linewidth]{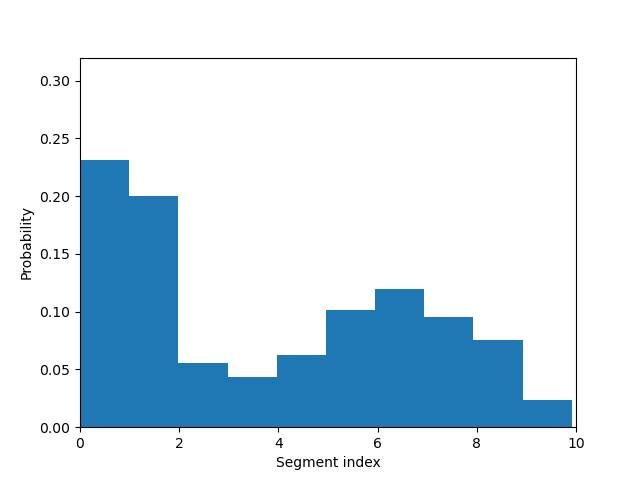}}
    \subfigure[GPT4-seg]{\label{fig:abc_prediseg}%
      \includegraphics[width=0.24\linewidth]{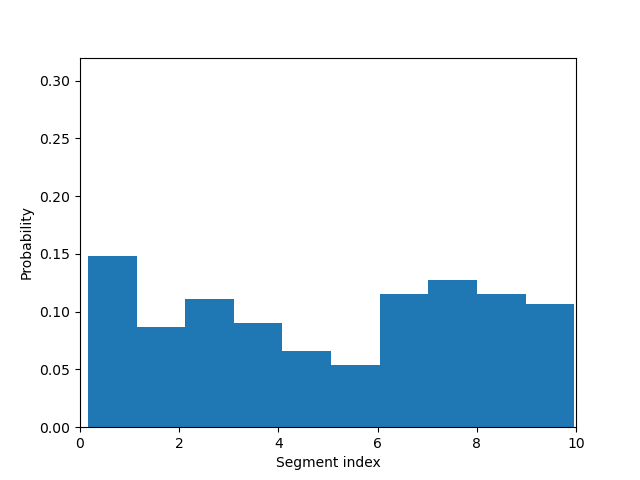}}
    \subfigure[GPT4-doc]{\label{fig:abc_predidoc}%
      \includegraphics[width=0.24\linewidth]{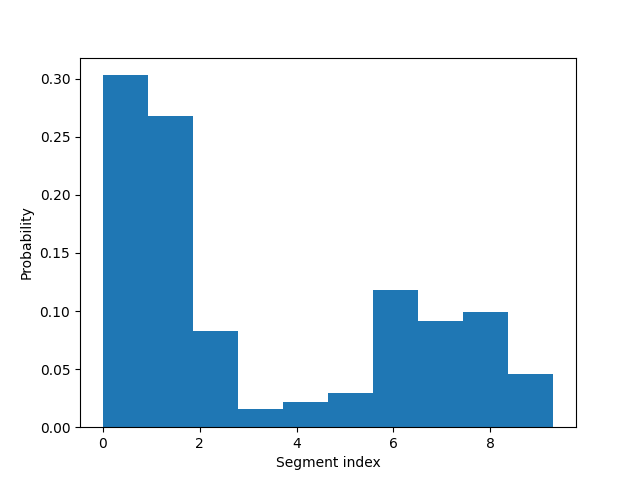}}
  }
\end{figure*}

\begin{figure*}[htbp]
\floatconts
  {fig:casestudy}
  {\caption{The precision-recall curve and PRAUC for two example diseases. Left: a) hypertension with a limited amount of missing mentions in the medical note; Right: b) anemia with many missing mentions. ModelA is the LLM Stage 1a with the \textit{Verifier} model, ModelB is the LLM Stage 1 with the \textit{Verifier} model and ModelC is LLM-codex with the LLM Stages 1 and 2 along with the \textit{Verifier} model.}}
  {\includegraphics[width=0.99\linewidth]{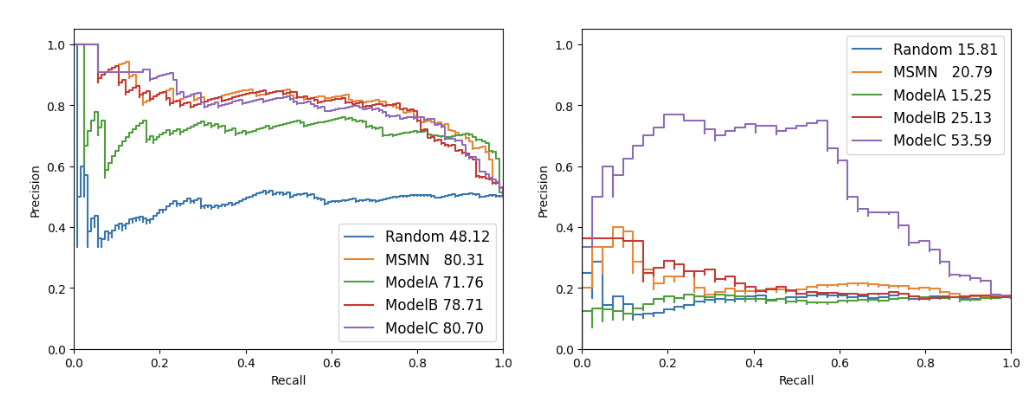}}
\end{figure*}
\comment{Joel: The figures will look nicer if you can render them as PDFs.}

\subsection{Prompt template}
\begin{example}
\label{example:icdcode_promptoriginal}
As a proficient clinical coding professionals, it is your responsibility to assign ICD 9 codes given the CLINICAL NOTE from the CANDIDATE LIST provided below.

    ---
    
    CLINICAL NOTE (or partial): 
    
    [text note]
    
    ---
    
    Here is a CANDIDATE LIST of 50 ICD 9 codes and their associated descriptions to assign: 
    
    [candidates]
    
    ---
    
For each disease/procedure based on the context in CLINICAL NOTE, you must generate a list of strings containing the ICD 9 codes you assigned.
  
\end{example}

\begin{example}
\label{example:evidence}
As a proficient clinical coding professional, it is your responsibility to extract evidence when assigning ICD code. Given the list of ICD 9 CANDIDATE codes (diseases/procedures) to assign, you need to verify each code by extracting associated evidence sentence from CLINICAL NOTE. 
You could inference based on basic medical commonsense, such as prescription of metformin is evidence to type 2 diabetes. 
 
 --- 
 
 ICD 9 CANDIDATE codes and descriptions: [diseases]. 
 
 ---
 
 Here is the CLINICAL NOTE split by sentence, each sentence starts with an index number surrounded by parentheses: [text note]
 
 --- 
 
 When assigning ICD code, you should: 
 
 1. Carefully assign ICD code to each sentence as evidence even ICD code is already assigned in the previous sentence; 
 
 2. If multiple ICD code found in one sentence, label them all and separate them by semicolon;
 
 3. Do not assign ICD code if it is negated or ruled out in the CLINICAL NOTE, for example you should not assign  "287.5" if "No leukemia or thrombocytopenia"; 
 
 4. Include ICD code only, not the associated English description. 
 
\end{example}

\end{document}